\documentclass[10pt, a4paper]{article}
\usepackage{booktabs}
\usepackage{bm}
\usepackage{amsthm}
\usepackage{amsmath}
\usepackage{mathrsfs}
\usepackage{graphicx}
\usepackage{authblk}
\usepackage[dvipdfm]{geometry}
\usepackage{arydshln}
\usepackage{times}
\usepackage[justification=centering]{caption}
\usepackage[figuresright]{rotating}
\usepackage{float}
\usepackage{url}
\usepackage[colorlinks,linkcolor=blue,citecolor=red,anchorcolor=blue]{hyperref}
\makeatletter

\long\def\pprintMaketitle{\clearpage
  \iflongmktitle\if@twocolumn\let\columnwidth=\textwidth\fi\fi
  \resetTitleCounters
  \def\baselinestretch{1}%
  \printFirstPageNotes
  \begin{center}%
 \thispagestyle{pprintTitle}%
   \def\baselinestretch{1}%
    \Large\@title\par\vskip18pt
    \normalsize\elsauthors\par\vskip10pt
    \footnotesize\itshape\elsaddress\par\vskip36pt
    \hrule\vskip12pt
    \ifvoid\absbox\else\unvbox\absbox\par\vskip10pt\fi
    \ifvoid\keybox\else\unvbox\keybox\par\vskip10pt\fi
    \hrule\vskip12pt
    \end{center}%
  \gdef\thefootnote{\arabic{footnote}}%
  }

\def\printFirstPageNotes{%
  \iflongmktitle
   \let\columnwidth=\textwidth\fi
  \ifx\@tnotes\@empty\else\@tnotes\fi
  \ifx\@nonumnotes\@empty\else\@nonumnotes\fi
  \ifx\@cornotes\@empty\else\@cornotes\fi
  \ifx\@elseads\@empty\relax\else
   \let\thefootnote\relax
   \footnotetext{\ifnum\theead=1\relax
      \textit{Email address:\space}\else
      \textit{Email addresses:\space}\fi
     \@elseads}\fi
  \ifx\@elsuads\@empty\relax\else
   \let\thefootnote\relax
   \footnotetext{\textit{URL:\space}%
     \@elsuads}\fi
  \ifx\@fnotes\@empty\else\@fnotes\fi
  \iflongmktitle\if@twocolumn
   \let\columnwidth=\Columnwidth\fi\fi
}

\long\def\@@address[#1]#2{\g@addto@macro\elsaddress{%
    \def\baselinestretch{1}%
     \refstepcounter{affn}
     \xdef\@currentlabel{\theaffn}
     \elsLabel{#1}%
    \textsuperscript{\theaffn}#2\par}}

\long\def\@address#1{\g@addto@macro\elsauthors{%
    \def\baselinestretch{1}%
    \addsep\footnotesize\itshape#1\def\addsep{\par\vskip6pt}%
    \def\authorsep{\par\vskip8pt}}}

\renewcommand\section{\@startsection {section}{1}{\z@}%
           {12\p@ \@plus 6\p@ \@minus 3\p@}%
           {9\p@ \@plus 6\p@ \@minus 3\p@}%
           {\normalsize\bfseries\boldmath}}

\renewcommand\subsection{\@startsection
 {subsection}{2}{0mm}
 {-\baselineskip}
 {0.5\baselineskip}
 {\normalfont\normalsize\itshape\bfseries}}

\renewcommand\subsubsection{\@startsection
 {subsubsection}{2}{0mm}
 {-\baselineskip}
 {0.5\baselineskip}
 {\normalfont\normalsize\itshape\bfseries}}

\makeatother

\begin{document}

\newtheorem{definition}{Definition}
\newtheorem{theorem}{Theorem}
\newtheorem{remark}{Remark}

\title{\bf{A brief introduction to the Grey Machine Learning}\footnote{This paper was awarded the Excellent Paper by GSUA in 2018 at Nanjing, China, and recommended to be published in Journal of Grey System after a second round of peer review.  To cite this paper: ``Ma, Xin. A brief introduction to the Grey Machine Learning. Journal of Grey System, 2019, vol. 31, no. 1, pp. 1-12."}}      
\author[1,2,\footnote{Correspondence author: Xin Ma, Email:cauchy7203@gmail.com}]{Xin Ma}

\date{}
\affil[1]{School of Science, Southwest University of Science and Technology, Mianyang, China}
\affil[2]{State Key Laboratory of Oil and Gas Reservoir Geology and Exploitation,
                  Southwest Petroleum University, Chengdu, China}

\maketitle%
\abstract{ This paper presents a brief introduction to the key points of the Grey Machine Learning (GML) based on kernels. The general formulations of the grey system models have been firstly summarized, and then the nonlinear extensions with general formulations of the grey models have also been developed . The kernel implicit mapping is used to estimate the unknown nonlinear function of the GML model, by extending the nonparametric formulation of the Least Squares Support Vector Machines (LSSVM), the estimation of the nonlinear function of the GML model can also be expressed by the kernels. A short discussions on the priority of this new framework to the existing grey models and LSSVM have also been presented. The open issues have been noted in detail.Perspectives and future orientations of this framework have also been presented.

}

\section{Introduction}

    The Artificial Intelligence (AI) has long be a hot spot, and recent breakthroughs have significantly raised the interest of the AI in recent few years. For instances,  Brenden Lake , $et \quad al$ \cite{handwrite} in 2015 designed a machine to learn to write the Mongolian, and it has passed the Turing test. This work proved that the machine can be act like humans in hand writing tasks. Another important breakthrough is the AlphaGo by Google DeepMind. David Silver, $et \quad al$ \cite{alphago} in 2016 published the first paper of AlphaGo and claimed that the computer can defeat the professionals of the game of Go for the first time, and actually this paper was published after the AlphaGo defeated the world champion Lee Se-dol. The success of AlphaGo further proved that the machine can not only be alike humans, but can also be smarter than humans. The success of these work brings unprecedented confidence to AI.

    The goal of AI is to build the machines which have intelligence like humans. The mathematical models and algorithms are essential to achieve this goal. In the works mentioned above, the Bayesian Framework has been used by Brenden Lake for the hand writing tasks, and the AlphaGo is actually built on the neural networks and Monte Carto tree search. However, these works also have another issues that they all need large scale of training data to obtain the optimal models. It is also very interesting to see that Google has developed another machine for playing the game of Go which is called the AlphaGo Zero \cite{alphagozero}. This new machine was trained without any chess manuals, only the rules of the game of Go was taught. The AlphaGo Zero has successfully won the champion-defeating AlphaGo \cite{alphago} by 100-0. In our opinion, this work also proved that it is possible to train the machines using extremely small samples. This also mentioned us that the Big Data should not be the only way for efficient AI.

    Actually, in the pattern of human learning we often encountered the cases with very small data. For example, a little child may master the basic tricks of the game of Go with a few lessons. Technically, the main methodology employed to build the AlphaGo is the Deep Learning and Reinforcement Learning, which are all based on the idea to take advantage of human knowledge as much as possible. The variations of the game of Go are actually computationally infinity for the computers. Thus, it is impossible to train a smart AlphaGo without the methods with efficient learning based on small samples.

    The grey system theory is primarily developed for the small samples. The grey models often appear to have deterministic structure with free parameters, which should be estimated by the samples. In the idea of the systems science, the deterministic structure of the grey models is essentially the known part the system, with the free parameters as the unknown part. In our opinion, such pattern is quite limited, as in the real world applications it is very difficult to discover the structure of the existing systems. And in most cases, we can only find out some of the part of the systems in a short time or by a low cost. Thus it is natural to ask the question: How can we simulate the systems with partially known structure ?

    Based on this question, we have carried out a series of works to combine the grey modelling techniques and the machine learning methods. The main problems considered in these works are the systems with known dynamical structure and the unknown nonlinear relationship between the output and input series. And the previous results in the real world applications all indicate that the models built on this idea are significantly more effective than the conventional grey models.

    In this paper, we would give the general formulation of this framework for the first time, and we name it as the Grey Machine Learning (GML). In order to better express the essence of this new framework, we abstracted the computational details for the general formulations in this paper. And for easier understanding and more effective applications, we tried our best to omit the deep mathematical theorems and concepts.

    The rest of this paper is organized as follows: In order to better explain the main idea of the GML, we firstly summarize the general forms of the conventional first order grey models in Section 2; Then the general formulation including the computational details are presented in Section 3; Short discussions on the main findings and open issues are shown in Section 4, and the perspectives are illustrated in Section 5.

\section{The typical formulation of the conventional grey models}

    Most of the existing grey models with continuous whitening equation share a general linear formulation as:
    \begin{equation}\label{eq:lincongreymodel}
       \frac{dX_1^{(1)}(t)}{dt} +a X_1^{(1)}(t)  = f(\bm{\theta};t),
    \end{equation}
    where the series $X_1^{(1)}$ is often called the feature series (or output series). The function $f(\bm{\theta};t)$ is often varied by time $t$ or reliance series (or input series) $X_i^{(1)},i=2,3,...,n$, with unknown parameters $\bm{\theta}$.

    For the discrete grey models, we can also write the general linear formulation as
    \begin{equation}\label{eq:lindisgreymodel}
        X_1^{(1)}(k+1) = \alpha X_1^{(1)}(k) + f(\bm{\theta};k).
    \end{equation}

    The notations have the same meaning to the continuous formulation \eqref{eq:lincongreymodel}.

    Most of the existing grey models can be written in the above formulations, and some popular ones with 1-AGO are listed in Table \ref{t:unimodels} and \ref{t:multimodels}. Moreover, many grey models without 1-AGO also follow such formulations. The grey model with fractional order accumulation by Wu et al. \cite{Wu2013GreyCNSNS} is one of the most popular grey models without 1-AGO, and this model still satisfies the formulation if we change the order of accumulation to be $r$. And similar cases can also be found in other references related to fractional grey models \cite{wu2018,zeng2017self,duan2018forecasting}. And it is also interested to see that some nonlinear grey models can also be transformed to such formulations, e.g. the nonlinear Bernoulli grey models \cite{maxin2019}.

    And it should be noticed that some nonlinear models can also be transformed into a linear form with similar formulations \eqref{eq:lincongreymodel} or \eqref{eq:lindisgreymodel}.

    \begin{table}[!htb]
    \centering
    \scriptsize
    \caption{Formulations of the existing univariate grey models\label{t:unimodels}}
    \begin{tabular}{cccc}
    \toprule
    Name	&	Model	&	$f(\cdot)$	&	References		\\
    \hline
    \\
     GM(1, 1)	&	$ \frac{dX^{(1)}(t)}{dt} + a X^{(1)}(t)  = f(\bm{\theta};t) $	&	$b$	&	\cite{liubook}\\
    \\
     NGM(1, 1, $k$)	&	$ \frac{dX^{(1)}(t)}{dt} + a X^{(1)}(t)  = f(\bm{\theta};t)$	&	$bt$	&	\cite{cui2013}	\\
    \\
     NGM(1, 1, $k$, $c$)	&	$ \frac{dX^{(1)}(t)}{dt} + a X^{(1)}(t)  = f(\bm{\theta};t) $	&	$bt+c$	&	\cite{zhanngm}\\
    \\
     TDPGM	&	$ \frac{dX^{(1)}(t)}{dt} + a X^{(1)}(t)  = f(\bm{\theta};t) $	&	$\sum_{\tau=1}^{t}[b\tau^2 + c\tau] +d $	&	\cite{maxintdpgm} \\
    \\
     DGM(1, 1) & $X^{(1)}(k+1) = \alpha X^{(1)}(k) + f(\bm{\theta};k)$ &  $\beta$ & \cite{xie2009}\\
    \\
     NDGM & $X^{(1)}(k+1) = \alpha X^{(1)}(k) + f(\bm{\theta};k)$ &  $\beta_1 k + \beta_2$ & \cite{xie2013}\\
    \bottomrule
    \end{tabular}
    \end{table}

    \begin{table}[!htb]
    \centering
    \scriptsize
    \caption{Formulations of the existing multivariate grey models\label{t:multimodels}}
    \begin{tabular}{cccc}
    \toprule
    Name	&	Model	&	$f(\cdot)$ &	Reference		\\
    \hline
    \\
    GM(1, $n$)	&	$ \frac{dX_1^{(1)}(t)}{dt} + a X_1^{(1)}(t)  = f(\bm{\theta};t) $	&	$\sum_{i=2}^{n}b_i X_i^{(1)}(t)$	&	\cite{liubook}\\
    \\
    GMC(1, $n$)	&	$ \frac{dX_1^{(1)}(rp+t)}{t} + b_1 X_1^{(1)}(rp+t)  = f(\bm{\theta};t) $	&	$\sum_{i=2}^{n}b_i X_i^{(1)}(t)+u$	&	\cite{tien2005}\\
    \\
    NGMC(1, $n$) & $\frac{dX_1^{(1)}(rp+t)}{t} + b_1 X_1^{(1)}(rp+t)  = f(\bm{\theta};t)$ &  $\sum_{i=2}^{n}b_i [X_i^{(1)}(t)]^\gamma+u$ & \cite{wangngmc2014}\\
    \\
    RDGM(1, $n$)	&	$ X_1^{(1)}(rp+k+1) = \beta_1 X_1^{(1)}(rp+k)  + f(\bm{\theta};k) $	&	$  \sum_{i=2}^{n}\beta_i Z_i^{(1)}(k)+\mu$	&	\cite{maxinrdgm1n}	\\
    \\
    DGM(1, $n$)	&	$ X_1^{(1)}(k+1) = \beta_1 X_1^{(1)}(k)  + f(\bm{\theta};k) $	&	$\sum_{i=2}^{n}\beta_i X_i^{(1)}(k)+\mu$	&	\cite{maxindgm1n}\\
    \\
    OGM(1, $n$) & $X_1^{(0)}(k) + a Z_1^{(1)}(k)  = f(\bm{\theta};k)$ &  $\sum_{i=2}^{n}b_i X_i^{(1)}(k)+h_1 \cdot (k-1) + h_2$ & \cite{zengboogm1n}\\
    \\
    NGM(1, $n$) & $X_1^{(0)}(k) + a Z_1^{(1)}(k)  = f(\bm{\theta};k)$ &  $\sum_{i=2}^{n}b_i [X_i^{(1)}(k)]^\gamma$ & \cite{wangngm2017}
    \\
    \bottomrule
    \end{tabular}
    \end{table}

    It is obvious that the solutions of the grey models in the above formulations also share the similar formulations. As is well known, most grey models use the initial condition $X_1^{(1)}(1)=X_1^{(0)}(1)$, then we can easily obtain the general formulation of such grey models.

    For the continuous models, the solution can always be presented as the following function with a convolution as:
    \begin{equation}\label{eq:conresponse}
       \hat{X}_1^{(1)}(t) = X_1^{(0)}(1)\cdot e^{-a(t-1)} + \int^{t}_{1} e^{-a(t- \tau )} f(\bm{\theta};\tau ) d \tau.
    \end{equation}

    For the discrete models, the solution can always be presented as the following function with a discrete convolution as:
    \begin{equation}\label{eq:disresponse}
       \hat{X}_1^{(1)}(k+1) = X_1^{(0)}(1)\cdot \alpha^{k} + \sum^{k+1}_{\tau=2} \alpha^{(k+1- \tau )} f(\bm{\theta};\tau ) .
    \end{equation}

    As shown above, the main structure is often known in a given grey model. The free parameters $a$ (or $\alpha$) and $\bm{\theta}$ are actually the unknown part of the grey models. Actually the structure of $f(\bm{\theta};t)$ (or $f(\bm{\theta};k)$) often plays a very important role in improving the grey models. $e.g.$ it is well known that the nonhomogeneous grey model NGM often performs much better than the GM(1, 1) model, and similar examples can be widely seen in the existing studies. But, we can not always find out an optimal formulation of the function $f(\cdot)$ in the real world applications. And this problem leads to our motivation of introducing the Machine Learning to the grey models.

\section{The Grey Machine Learning}

\subsection{The General formulation}
   The general formulation of the GML models share the equations as:

   The continuous form:
    \begin{equation}\label{eq:congreymodel}
       \frac{dX_1^{(1)}(t)}{dt} +a X_1^{(1)}(t)  = \phi(t) + u,
    \end{equation}

   The discrete form:
    \begin{equation}\label{eq:disgreymodel}
        X_1^{(1)}(k+1) = \alpha X_1^{(1)}(k) + \phi(k) + \mu.
    \end{equation}

   Where the function $\phi(\cdot)$ is totally unknown, and this is also the main difference between the GML models and the conventional grey models.

\subsection{Estimation of the unknown function $\phi(\cdot)$}

    Now we need to estimate the unknown function $\phi(\cdot)$ in the general formulations of the GML models. The kernel method has been mainly used in our previous works. In this subsection, we will discuss it in a more concise way.

\subsubsection{Linear representation in a higher dimensional feature space}

    According to the Weierstrass Theorem, any continuous function on a closed and bounded interval can be uniformly approximated by polynomials to any degree of accuracy. Less formally, we can write the approximation of any function in the following form
    \begin{equation}\label{eq:phi1}
        \phi(t) = w_n x^{n}(t) + w_{n-1} x^{n-1}(t)+...+w_1 x^{1}(t) + w_0.
    \end{equation}

    Let the mapping be $\varphi: R \rightarrow \mathscr{F}$, where $\mathscr{F}= \{[x^{n}(t), x^{n-1}(t),...,x^{1}(t),1]^T\}$, and the weights be $\omega=[w_n,w_{n-1},...,w_1,w_0]^T$. Then the nonlinear function can be written in a linear form in the higher dimensional feature space $\mathscr{F}$ as
    \begin{equation}\label{eq:phi2}
        \phi(t) = \omega^T \varphi(x(t)).
    \end{equation}

    Notice that this formulation always holds even the $x(t)$ is a vector. And the dimension of the feature space $\mathscr{F}$ can be infinity.

\subsubsection{The nonparametric estimation}

    Within the discussion above, we can transform the estimation of arbitrary nonlinear function into a linear problem. With the given samples $\{(x(k),y(k))|k=1,2,3,...,N\}$, we want to have the estimation in the following formulation
    \begin{equation}\label{eq:y1}
        y(t) = \omega^T \varphi(x(t)) + b.
    \end{equation}

    One defines the regularization problem, which is also in the form of ridge regression, as follows:
    \begin{equation}\label{eq:ridge}
    \begin{aligned}
            \min & J(\omega,\bm{e})  = \frac{||\omega||^2}{2} + \frac{C}{2} \sum_{k=1}^{m} e_k^2 \\
            s.t.\quad& e_k = y(k) -  \omega^T  \varphi (x(k)) - b,
    \end{aligned}
    \end{equation}
    where $||\cdot||$ is the 2-norm. The main difference between the regularization problem \eqref{eq:ridge} and the commonly used least squares method is that the  $||\omega||^2$ is also expected to be minimum. Actually, the available nonlinear mapping $\varphi$ is not unique. e.g. if the nonlinear function $\phi(t)$ is differentiable with higher order, it can be expanded with the Taylor power series, and there are numerous formulations if $\phi(t)$ is expanded at different points. Thus, mathematically the ridge regression is used to ensure the solution to be unique. And also, with nondeterministic mathematical expressions, the regularization problem is also categorized in the nonparametric estimation formulations. On the other hand, we often want the estimation to be plat enough to have higher generalization, or more stable. To this end, the term $C$ should be used.

    The regularization problem \eqref{eq:ridge} is essentially a constrained quadric programming, thus we need firstly to define the Lagrangian as

    \begin{equation}\label{eq:lag}
            L:=  \frac{||\omega||^2}{2} + \frac{C}{2} \sum_{k=1}^{r} e_k^2 +\sum_{k=1}^{m} \lambda_k  [ y(k) - \omega^T  \phi (x (k)) -  b  - e_k ],
    \end{equation}

    The solution of the regularization problem \eqref{eq:ridge} can be easily obtained using the KKT conditions as follows:
    \begin{equation}\label{eq:kkt}
           \left\{
           \begin{aligned}
              \frac{\partial L}{\partial w}=0    &\Rightarrow  \omega =  \sum_{k=1}^{m} \lambda_k \varphi (x(k)) \\
              \frac{\partial L}{\partial b}=0    &\Rightarrow  \sum_{k=1}^{m} \lambda_k=0  \\
              \frac{\partial L}{\partial e_k}=0  &\Rightarrow  e_k= \lambda_k/C \\
              \frac{\partial L}{\partial \lambda_j}=0  &\Rightarrow  y(k)  - \omega^T  \varphi (\chi (k)) -  b  = e_k.
           \end{aligned}
            \right.
    \end{equation}

    With the first equation in \eqref{eq:kkt}, we can see that the $\omega$ can be computed by the values of the Lagrangian multipliers $\lambda_k$ and $\varphi (x(k))$. Thus we can easily obtain the computational formulation of the nonlinear function as
    \begin{equation}\label{eq:y2}
         \omega^T \varphi(x(t)) =  \sum_{k=1}^{m} \lambda_k \varphi^T (x(k))\varphi (x(t)).
    \end{equation}

    We have not give the deterministic form of the nonlinear mapping $\varphi (\cdot)$ yet. But notice that we only need to compute the inner product $<\varphi(x(k)),\varphi (x(t))>=\varphi^T (x(k))\varphi (x(t))$ in the feature space $\mathscr{F}$. This is quite simple if we use the Mercer's conditions (See [31]), with which we know that for any symmetric positive definite function $K(\cdot,\cdot)$, we have the following expansion:

    \begin{equation}\label{eq:mercer}
         K(x(k),x(t)) =  \sum_{n=1}^{\infty} \alpha_n \psi_n (x(k))\psi_n (x(t)),
    \end{equation}
    where $\{\psi_n (x(t))\}_{n=1}^{n=\infty}$ is the orthogonal basis, and $\alpha_n$ are the nonnegative eigenvalues. And such functions are often called the kernel functions or kernels.

    Thus we can easily define the nonlinear mapping corresponding to a given kernel as
    \begin{equation}\label{eq:phikernel}
         \varphi (\cdot) = [\sqrt{\alpha_1} \psi_1 (\cdot),\sqrt{\alpha_2} \psi_2 (\cdot),...,\sqrt{\alpha_n} \psi_n (\cdot),...]^T.
    \end{equation}

    And obviously, the inner product of the nonlinear mapping can be written as the kernel
    \begin{equation}\label{eq:inner}
    \begin{aligned}
         K(x(k),x(t)) &=  \sum_{n=1}^{\infty} \alpha_n \psi_n (x(k))\psi_n (x(t))\\
         &=<[\sqrt{\alpha_1} \psi_1 (k),\sqrt{\alpha_2} \psi_2 (k),...,\sqrt{\alpha_n} \psi_n (k),...]^T,[\sqrt{\alpha_1} \psi_1 (t),\sqrt{\alpha_2} \psi_2 (t),...,\sqrt{\alpha_n} \psi_n (t),...]^T>\\
         &= \varphi^T (x(k))\varphi(x(t)).
    \end{aligned}
    \end{equation}

    Thus substituting \eqref{eq:y2}, \eqref{eq:inner} into \eqref{eq:kkt} with eliminating the $e_k$, the solution of the regularized problem shares the following linear system
    \begin{equation}\label{eq:l1}
            \left( \begin{array}{c;{2pt/2pt}c}

                        0 & \bm{1}^{T}_{m}   \\
                       \hdashline[2pt/2pt] \\
                       \bm{1}_{m}  & \Omega +  I_{m}/C
                    \end{array}
            \right)
            \left( \begin{array}{c}
                       b \\ \hdashline[2pt/2pt] \\ \bm{\lambda}
                    \end{array}
            \right)
            =
            \left( \begin{array}{c}
                     0 \\ \hdashline[2pt/2pt] \\ Y
                    \end{array}
            \right)
        \end{equation}
    where
    $$ \bm{1}_{m}=[1,1,...,1]^T_{m} ,$$
    $$ \Omega= \left[ K(x(i),x(j)) \right]_{m \times m} ,$$
    $$ \bm{\lambda}=[\lambda_1,\lambda_2,\lambda_3,...,\lambda_m]^T ,$$
    $$ Y=[y(1),y(2),...,y(m)]^T,$$
    and $I_{m}$ is the identity matrix.

    Now we can see that the solution of the regularized problem \eqref{eq:ridge} has been converted to the linear system \eqref{eq:l1}, which can be easily solved with a given kernel, and the results are the values of $\bm{\lambda}$ and $b$.

    At the last stage, we need to give the computational formulation of the estimation \eqref{eq:y1}. Within the \eqref{eq:y2} and \eqref{eq:inner}, the estimation of \eqref{eq:y1} can be easily obtained as
    \begin{equation}\label{eq:inner}
    \begin{aligned}
            \hat{y}(t) & = \omega^T \varphi(x(t)) + b \\
                 & = \sum_{k=1}^{m} \lambda_k \varphi^T (x(k))\varphi (x(t)) + b \\
                 & = \sum_{k=1}^{m} \lambda_k K(x(k),x(t)) + b.
    \end{aligned}
    \end{equation}

\subsubsection{The semiparametric estimation}

    With the nonparametric estimation, we can easily deduce the semiparametric version, of which the objective is to obtain the estimation shares the following formulation
    \begin{equation}\label{eq:ypartial}
        y(t) = \omega_{1}^T z(t) +  \omega_{2}^{T} \varphi(x(t)) + b,
    \end{equation}
    where $z(t)$ is a variable in $R$ or a vector in the linear space $R^{N_1}$. This form can be easily transformed to the nonparametric formulation by denoting a new nonlinear mapping $\varphi ': R^{N_1 + N_2} \rightarrow \mathscr{F}'$ as
    \begin{equation}\label{eq:phikernel2}
         \varphi' (\cdot) =
       \left[ \begin{aligned}
             z(\cdot) \\ \varphi (\cdot)
        \end{aligned}\right],
    \end{equation}
    and set $\omega'=\left[ \begin{aligned}
             \omega_1 \\ \omega_2
        \end{aligned}\right],$

    The Eq.\eqref{eq:ypartial} can be written as
    \begin{equation}\label{eq:ypartial2}
        y(t) =   {{\omega}'}^{T} \varphi'(x(t)) + b.
    \end{equation}

    Notice that this formulation is mathematical equivalent to the nonparametric form \eqref{eq:y1}, thus all the computational formulations are just the same. And now we can simply obtain the key formulations for the semiparametric estimation with some tiny changes.

    Firstly, the $\Omega$ in \eqref{eq:kkt} can be rewritten as
    \begin{equation}\label{eq:omega2}
    \begin{aligned}
        \Omega_{ij} &=   <\varphi'(x(i)),\varphi'(x(j))>\\
                    &=  [z^T(i),\varphi^T (x(i))]\left[ \begin{aligned}[l]
             z(j) \\ \varphi (x(j))
        \end{aligned}\right]\\
                    &= z^T(i)z(j) + \varphi^T (x(i))\varphi (x(j))\\
                    &= z^T(i)z(j) + K(x(i),x(j)).
    \end{aligned}
    \end{equation}

    Being similar to \eqref{eq:y2}, the nonparametric form \eqref{eq:ypartial2} can also be transformed to
    \begin{equation}\label{eq:ypartial3}
    \begin{aligned}
        \hat{y}(t) & =   {{\omega}'}^{T} \varphi'(x(t)) + b\\
             & = \sum_{k=1}^{m} \lambda_k {{\varphi}'}^T (x(k)){{\varphi}'} (x(t))\\
             & = \sum_{k=1}^{m} \lambda_k z^T(k)z(t) + \sum_{k=1}^{m} \lambda_k K(x(k),x(t)) + b.
    \end{aligned}
    \end{equation}

    If we note $\omega_1=\sum_{k=1}^{m} \lambda_k z(k)$, we can also write \eqref{eq:ypartial3} as
    \begin{equation}\label{eq:ysemi}
        \hat{y}(t) =  \omega_1 ^T z(t) + \sum_{k=1}^{m} \lambda_k K(x(k),x(t)) + b.
    \end{equation}

    From the above discussions, we can see that the nonlinear function $\phi(t)$ is always estimated as the form with kernels:
    \begin{equation}\label{eq:ysemi}
        \phi(t) =  \sum_{k=1}^{m} \lambda_k K(x(k),x(t)).
    \end{equation}

    This is very important as it ensures the computational feasibility of the grey systems with an unknown nonlinear function, and within this it is quite simple to derive the estimation of the general forms \eqref{eq:congreymodel} and \eqref{eq:disgreymodel}. e.g. if we use the \eqref{eq:disgreymodel}, we can easily obtain the computational formulations by the following simple notations:
    \begin{equation}\label{eq:notegrey}
        \left\{\begin{aligned}
                            y(k+1)& = X_1^{(1)}(k+1) \\
                            \omega_1 &= \alpha \\
                            z(k) &= X_1^{(1)}(k) \\
                            \phi(k) &= \omega_2^T \varphi(\chi(k))\\
                            \chi(k) &= [X_2^{(1)}(k),X_3^{(1)}(k),...,X_n^{(1)}(k)]^T.
                \end{aligned}\right.
    \end{equation}

\subsection{The solutions of the general formulations}

    Mathematically, the solutions of the general formulations shares the similar expressions to the linear ones as follows:

    The continuous form of \eqref{eq:congreymodel}:
    \begin{equation}\label{eq:csol}
       \hat{X}_1^{(1)}(t) = X_1^{(0)}(1)\cdot e^{-a(t-1)} + \int^{t}_{1} e^{-a(t- \tau )} \phi(\tau ) d \tau.
    \end{equation}

    The discrete form of  \eqref{eq:disgreymodel}:
    \begin{equation}\label{eq:dsol}
       \hat{X}_1^{(1)}(k+1) = X_1^{(0)}(1)\cdot \alpha^{k} + \sum^{k+1}_{\tau=2} \alpha^{(k+1- \tau )} \phi(\tau ).
    \end{equation}

    Notice that we have only changed the linear function $f(\cdot)$ in \eqref{eq:conresponse} and \eqref{eq:disresponse} into $\phi(\cdot)$.

\subsection{Implementations}
    The details on how to deal with the continuous and discrete solutions can be found in our previous works
     \cite{krngm,kgm1n,ma2016mpe,krgmc1n}.

    The implementation of the GML models is also very simple. The well documented Matlab packages of the following GML models are available online. One can easily build particular GML models combining the general framework of GML and the source codes.

    \item{ The KGM(1, n) model\footnote{Source code \\ \url{http://cn.mathworks.com/matlabcentral/fileexchange/65221-the-kernel-based-grey-system-model}}}

    \begin{equation}\label{eq:kgm1n}
        X_1^{(0)}(k+1) + \alpha Z_1^{(1)}(k+1) = \phi(k) + \mu.
    \end{equation}

    \item{ The KNEA model \footnote{In the original paper the $X_1^{(0)}(k)$ is denoted as $q(k)$ as it stands for the oil production. \\ Source code \\ \url{https://ww2.mathworks.cn/matlabcentral/fileexchange/58918-the-kernel-regularized-extension-of-the-arps--decline-model--knea-}}}

    \begin{equation}\label{eq:knea}
        X_1^{(0)}(k+1) = \alpha X_1^{(0)}(k) + \phi(k) + \mu.
    \end{equation}

\section{Short discussions on the findings and issues}

\subsection{Main findings}
     Actually, the nonparametric estimation presented above is essentially the classical Least Squares Support Vector Machines (LSSVM) by Johan Suykens $et.al.$ \cite{lssvm}, which represents one of the typical framework of the Machine Learning methods. The semiparametric estimation is the variation of the Partially Linear Least Squares Support Machines (PL-LSSVM), which was proposed by Marcelo Espinoza, Johan Suykens and Bart De Moor in 2004 \cite{pllssvm}, which has not been paid much attention in the past years. However, it can be seen in this paper that this formulation is quite useful to build the GML models. In fact a very strong result called the Representer Theorem by Bernhard Sch{\"o}lkopf, $et.al.$ has been proved in 2001 \cite{representer2001}, which shows that for any regularized formulation of regularized problems for kernel approximation, the estimation always follows the form of \eqref{eq:ysemi}. The above facts show that the computational and theoretical basis of the kernel method is well established, which can make us more confident to use the kernel methods in the GML framework.

     As mentioned above, our objective is to build the real ``Grey" models with partially known part and partially unknown part. According to the general forms of the GML models represented above, the partially known part is actually the linear differential or difference in \eqref{eq:congreymodel} or \eqref{eq:disgreymodel}. Such formulation often implies that the status or output series of the systems is declining or increasing by time, and such systems are often called the dynamical systems, which widely exist in the real world applications, such as the oil \& gas fields, batteries, etc. The nonlinear function estimated by the kernels represents the nonlinear function of the input series or reliance series of the systems. The physical meaning of these functions are clear, which essentially describes the nonlinear relationship in the dynamical systems. Thus, the GML models essentially represent the nonlinear dynamical systems.

     With such properties, the GML models have been firstly shown to be more efficient than the existing linear grey models, these findings proved the nonlinearity of the new models. In our recent works \cite{kgm1n}, the GML models have also been shown to outperform the classical LSSVM, which is essentially a static model, and this proved the dynamical property of the GML models. What's more, the comparison to the LSSVM also proved that the use of the known information can significantly improve the performance of the conventional Machine Learning models. This is also very important to the existing Machine Learning methods.

\subsection{The open issues}

    The GML is essentially a combination of the existing grey modelling method and kernel method, thus it would also inherit the existing flaws of these methods.

    The main computational pattern actually follows the general formulation of the linear grey models. Thus the existing issues on the existing grey models may also effect the performance of the GML models. Such as the issues on background values, \cite{zb2018,zeng2018improved}), initial point optimization \cite{xie2009}, the inconsistent problems \cite{maxingmco}, $etc$. However, with  nonlinear formulations of the GML models, it is not clear whether the existing methods are still available at present.

    On the other hand, as the GML framework employs the kernel method, it was expected that some important properties of the existing kernel-based models would still exist, but it was not the way as shown in our works.

    (1)\emph{Selection of kernels} It was reported that the kernels could have the same performance with proper kernel parameters \cite{smolabook}. But it was found that only the Gaussian kernel can be efficient in the GML models \cite{krngm} in the numerical experiments, and the efficiency of the GML model with Gaussian kernel can be much higher than that with other kernels. This implies that we can not easily use the existing knowledge of the established kernel learning theories, and the existing results should be fixed for the GML models.

    (2)\emph{Optimization of hyperparameters} It is also well known that the regularized parameter (e.g. $C$ in \eqref{eq:ridge}) and the kernel parameters (the tunable parameters in the kernel functions), which are often called hyperparameters uniformly,  are also very important to the kernel-based models, and sometimes they are even more important than the formulation of the kernels. The most commonly used method for tuning such parameters is the cross validation (CV). However, the CV was shown to be available for only a few cases \cite{krngm,ma2016mpe}. Although some works have been presented to show the effectiveness of the CV for the time series models, there still exist controversies on such point of view. And what's more, as the GML is mainly designed for small samples, it would be more difficult for the theoretical analysis.

    (3)\emph{Training algorithm} There numerous training algorithms for the kernel-based models, but in this paper we have not mentioned it in the above content. The main reason is that the problems we are willing to solve by the GML are the ones with small samples, and the training task is never difficult. Out of interest, we still have carried out some researches, obtaining some interesting results. The main stream training algorithms for the kernel-based models can be simply categorized in two classes, the gradient based algorithms and the SMO like algorithms. The SMO are often reported to be the best choice in the applications for the commonly used kernel-based models, such as the standard SVM \cite{svmsmo} and LSSVM \cite{lssvmsmo}. However, in our experiments, the SMO can never outperform the conjugate gradient (CG) method \cite{maxinphd}. And the SMO needs millions of iterations even in some very simple data sets. In our opinion, although the CG is enough for us to train the GML models, it is still very interesting to figure out why the SMO loses its priorities in the GML. And the results may bring us some useful knowledge for the numerical computations.

\section{Conclusions and Perspectives}

     According to the above discussions, our idea of the ''Grey" models can be implemented using the semiparametric estimation by the kernels. With the general formulation of the GML, more efficient GML models can be developed in the future.

     Further, the works of GML at present do not only build some new grey models, but also prove the possibility to combine the dynamic nature of the grey system models and the nonlinearity of the Machine Learning models. Thus some other Machine Learning methods could also be expected to be used to build the GML models, such as the Multilayer Perceptrons (MP), Gaussian Process Regression (GPR), Deep Learning Neural Networks, $etc$.

     At last, as mentioned above, with the widely existence of the nonlinear dynamical systems, more works in a wider range of real world applications should also be carried out. Especially the unconventional oil and gas systems \cite{Hu2018streamlineSPEPO,Wang2018FlowIJNSNS}, building systems (energy consumption and emissions) \cite{Minda2019}, and clean energy systems \cite{wu2018,Du2018multistepRE,Fan2019,Fan2018}.

\section{Acknowledgment}
   This research was supported by the Open Fund (PLN201710) of  State Key Laboratory of Oil and Gas Reservoir Geology and Exploitation (Southwest Petroleum University), and the Doctoral Research Foundation of Southwest University of Science and Technology (no. 16zx7140).

\bibliographystyle{unsrt}
\bibliography{refs}
\end{document}